\documentclass[runningheads]{llncs}
\usepackage[T1]{fontenc}
\usepackage{graphicx}

\usepackage{booktabs,multirow}
\usepackage{xcolor}
\usepackage{amsmath}

\begin{document}
\title{Towards Explainability in Monocular Depth Estimation
}
%
%
\author{Vasileios Arampatzakis\inst{1,2}\orcidID{0000-0003-4320-3740} \and George Pavlidis\inst{2}\orcidID{0000-0002-9909-1584} \and Kyriakos Pantoglou\inst{2}\orcidID{0009-0008-5683-3382} \and Nikolaos Mitianoudis\inst{1}\orcidID{0000-0003-0898-6102} \and Nikos Papamarkos\inst{1}\orcidID{0000-0003-2730-0006}}
\authorrunning{Arampatzakis et al.}
%
\institute{Democritus University of Thrace, Xanthi, Greece\\
\email{\{vaarampa, nmitiano, papamark\}@ee.duth.gr}\\ 
\and
Athena Research Center, Xanthi, Greece\\
\email{\{vasilis.arampatzakis, kyriakos.pantoglou, gpavlid\}@uathenarc.gr}}
\maketitle              
\begin{abstract}

The estimation of depth in two-dimensional images has long been a challenging and extensively studied subject in computer vision. Recently, significant progress has been made with the emergence of Deep Learning-based approaches, which have proven highly successful. This paper focuses on the explainability in monocular depth estimation methods, in terms of how humans perceive depth. This preliminary study emphasizes on one of the most significant visual cues, the relative size, which is prominent in almost all viewed images. We designed a specific experiment to mimic the experiments in humans and have tested state-of-the-art methods to indirectly assess the explainability in the context defined. In addition, we observed that measuring the accuracy required further attention and a particular approach is proposed to this end. The results show that a mean accuracy of around 77\% across methods is achieved, with some of the methods performing markedly better, thus, indirectly revealing their corresponding potential to uncover monocular depth cues, like relative size.
\keywords{computer vision \and monocular depth estimation \and explainability.}
\end{abstract}
\section{Introduction}

Research by Nagata \cite{nagata1987reinforce}, further classified by Cutting and Vishton \cite{cutting1995perceiving}, presented a complete set of visual depth cues, as a result of specifically designed experiments in humans. These cues include \textit{occlusion}, \textit{relative size}, \textit{relative density}, \textit{height in the visual field}, \textit{aerial perspective}, \textit{motion perspective}, \textit{convergence}, \textit{accommodation}, and \textit{binocular disparity}. The combination of these visual depth cues appears to be associated with the comprehensive understanding of a scene. Later, other researchers tried to systematically review the domain and present a more thorough view of depth estimation in humans \cite{howard2012perceiving}. 

The scientific community has long faced a significant challenge in achieving depth perception in mechanical systems. The accurate estimation of depth is a crucial task in machine visual perception. Depth estimation involves reconstructing the missing dimension, which represents the distance between the objects and the observer in a three-dimensional (3D) scene, through a two-dimensional (2D) projection of the scene. 

Nevertheless, there is no study, to the best of our knowledge, that clearly connects depth cues, as defined for humans, with the ability of modern depth estimation methods to estimate depth in monocular 2D images, thus affecting the explainability of those methods. In this paper, we explore towards addressing this issue, by using specifically designed data, and focus on the relative size depth cue, a prominent cue in projected scenes. We evaluate selected state-of-the-art depth estimation methods using those data and provide insights into their explainability, within our context.

\section{Deep Learning-Based Monocular Depth Estimation}

Traditional methods relied on assumptions, constraints, and optimizations to provide detailed depth estimates. However, these methods faced limitations such as a restricted measurement range, sensitivity to outdoor lighting conditions, calibration requirements, and high energy consumption, which hindered the utilization of sensor-based techniques involving RGB-D and LiDAR sensors. Additionally, approaches based on image pairs or sequences could only calculate depth values for sparse points. 

To tackle these challenges, researchers proposed the usage of deep learning. Deep Learning methods achieved high performance in estimating dense depth maps. In tasks like depth calculation with high complexity, where it is nearly impossible to apply classical pattern recognition approaches, deep learning methods achieved remarkable results. In the following study we focus on the significant work by Ibraheem et al. \cite{alhashim2018high}, Godard et al. \cite{godard2019digging}, and Ranftl et al. \cite{ranftl2020towards}, who made important contributions to the field. Ibraheem et al. \cite{alhashim2018high} proposed a novel approach (\textbf{DenseDepth}) to estimate high-resolution depth maps from RGB images using a transfer learning-based encoder-decoder network. The encoder was a pretrained DenseNet-169, fine-tuned on NYU Depth v2\footnote{NYU-v2, indoor images (rooms \& hallways scenes) \cite{silberman2012indoor}.} and KITTI datasets\footnote{KITTI, outdoor images (urban \& street scenes) \cite{geiger2012we}.}. The authors reported state-of-the-art results in typical and qualitative aspects and in terms of generalisation. Godard et al. \cite{godard2019digging} focused on estimating depth using video sequences, stereo pairs, or a combination of both (\textbf{Monodepth2}). They introduced several improvements. The system was trained on subsets of the KITTI dataset. The authors reported state-of-the-art results, outperforming other methods at that period. Ranftl et al. \cite{ranftl2020towards} proposed a novel approach (\textbf{MiDaS}) to enhance the robustness of depth estimation models and address the challenge of dataset bias. The authors reported results outperforming previous methods and particularly in terms of generalisation, making their method one of the most effective to date.

\section{The explainability experiment and results}
\label{section:Experiment}

\begin{figure}[!t]
\centering
\begin{minipage}{0.23\linewidth}
\centering\includegraphics[width=1\textwidth]{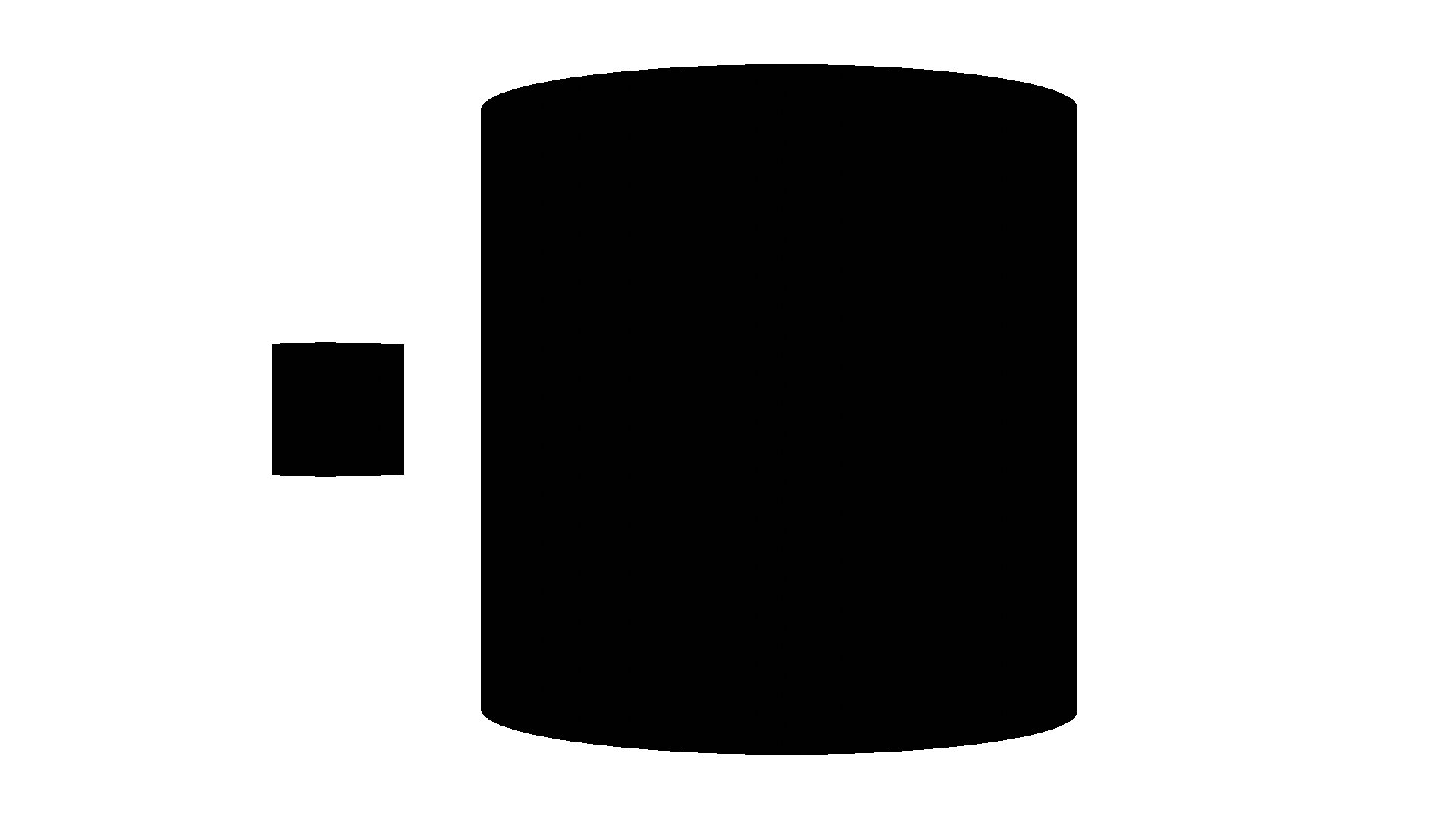}
\end{minipage}
\hfill
\begin{minipage}{0.23\linewidth}
\centering\includegraphics[width=1\textwidth]{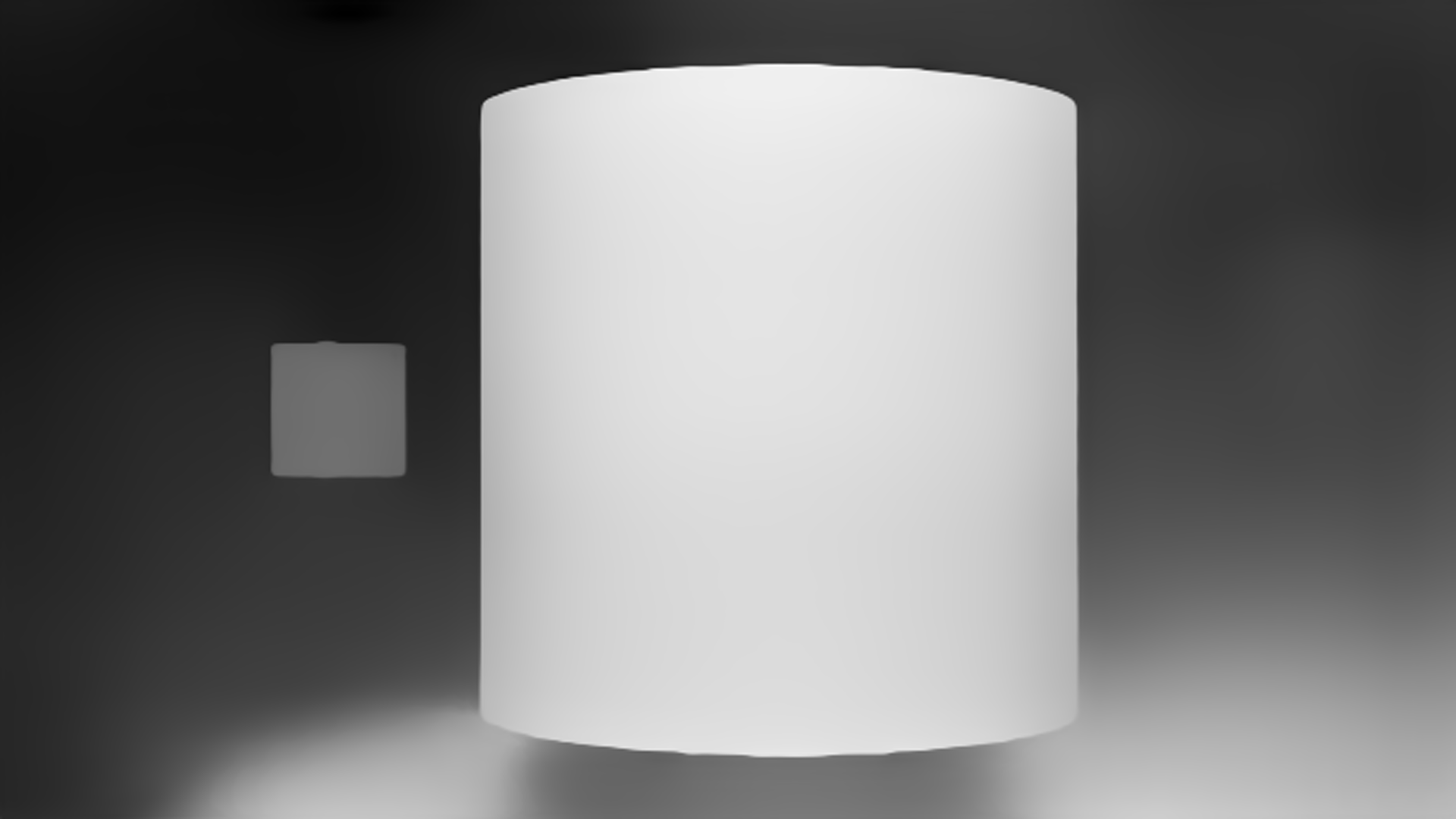}
\end{minipage}
\hfill
\begin{minipage}{0.23\linewidth}
\centering\includegraphics[width=1\textwidth]{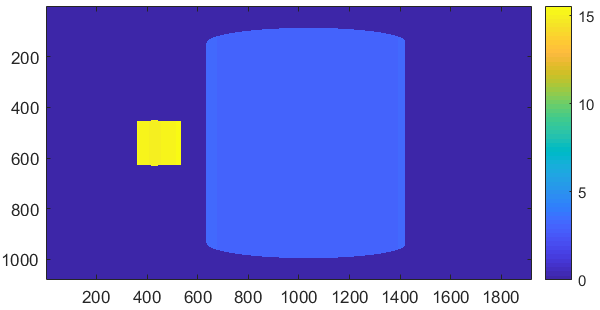}
\end{minipage}
\hfill
\begin{minipage}{0.23\linewidth}
\centering\includegraphics[width=1\textwidth]{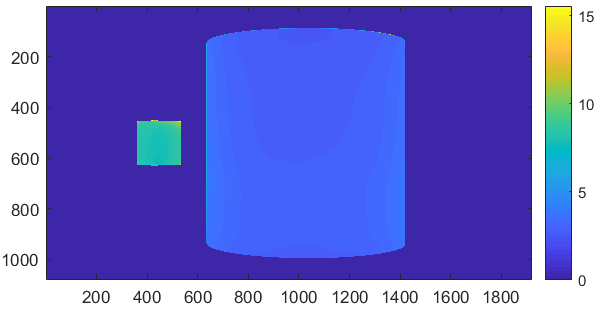}
\end{minipage}
\begin{minipage}{0.23\linewidth}
\centering (a)
\end{minipage}
\hfill
\begin{minipage}{0.23\linewidth}
\centering (b)
\end{minipage}
\hfill
\begin{minipage}{0.23\linewidth}
\centering (c)
\end{minipage}
\hfill
\begin{minipage}{0.23\linewidth}
\centering (d)
\end{minipage}
\caption{Process example: (a) original image, (b) predicted depth, (c) pseudo-coloured groundtruth depth, (d) pseudo-coloured masked predicted depth.}
\label{fig:example}
\end{figure}

To create meaningful explainability experiments we tried to mimic the relevant work done in humans. In the original experiments, as described in the Introduction, humans were asked to assess the relative distance of partially viewed untextured objects against neutral backgrounds. Thus, a particular artificial dataset had to be created, using 3D modeling software. The new dataset consists of 23800 2D images of black cylindrical objects at various distances against a white background, created through perspective projections of the corresponding virtual 3D scenes. Those data were used to test the three selected pretrained state-of-the-art methods. The core idea is to indirectly assess the explainability of the methods in learning the relative size cue, by providing test examples which contain only this single cue. We considered all published variations of the considered models. Furthermore, to evaluate the models, binary masks were applied to focus only on the pixels associated with objects in the scene. In addition, we adopted the scale and shift pre-alignment suggested by \cite{ranftl2020towards}. The assessment of depth predictions was based on common error and accuracy metrics, the most popular of which were introduced by Eigen \cite{eigen2014depth}; we used those definitions. An indicative example of the process is shown in \figurename~\ref{fig:example}, where an original image of two objects at different distance is shown in (a), the predicted depth in (b), and pseudo-coloured representations of the groundtruth (c) and the predicted depth (d). 

\begin{figure}[!t]
\centering
\includegraphics[width=\linewidth]{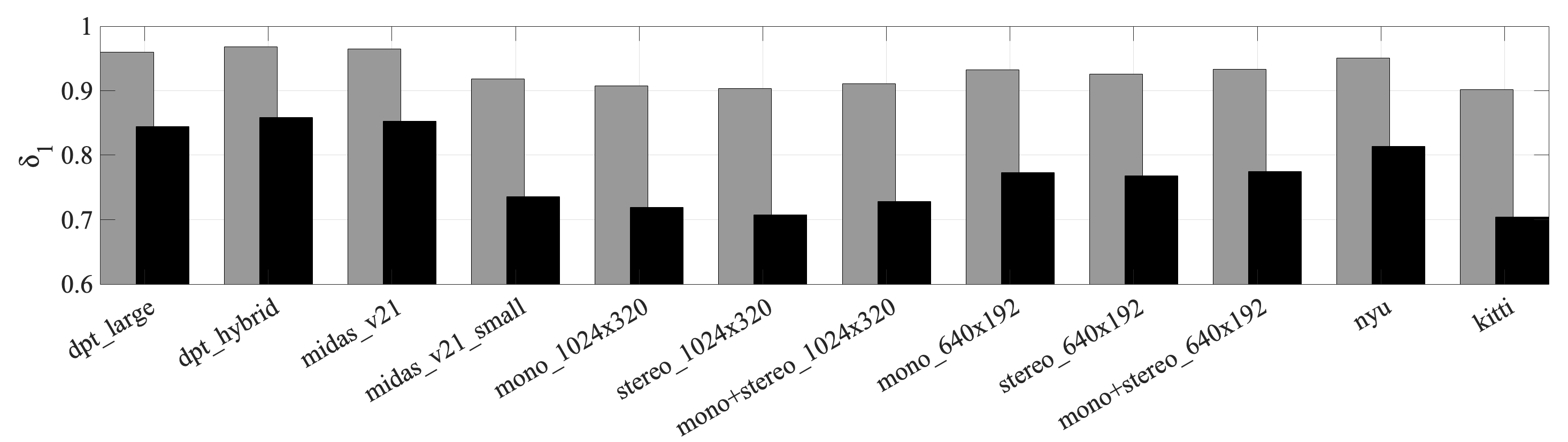}
\caption{The accuracy with a threshold $\delta_1$ of the pertained models tested on our dataset.}
\label{fig:relative_size_results}
\end{figure}

Additionally, we observed that the size of the objects in the images plays a significant role in the estimates of the error and accuracy. In the example shown, an error in the estimated depth of the far object (depicted significantly smaller) will have a negligible impact on the metrics, while the estimate should be balanced. This observation led to the introduction. After the objects become equally sized, the metrics are calculated, ensuring equal significance in error estimates for both objects in the scene. \figurename~\ref{fig:relative_size_results} depicts the overall average results. The gray bars represent results obtained by using the metrics in the typical way, whereas the black bars represent results obtained after the rescaling process. As expected, rescaling the smaller object results in lowering the accuracy (increasing the error) on the average.

Apparently, the three first variations of MiDaS outperform the other methods, particularly when using rescaling, and achieve an average of $\delta_1=0.85$ (85\%). The same is reflected in the error estimates. From the experiments, it seems possible that MiDaS partially learns the relative size cue, although more experiments are needed to generalise this remark. In addition, the DenseDepth method pretrained on NYU-v2 also exhibits increased accuracy compared with MiDaS, whereas the same model pretrained on KITTI is among the worst cases. This shows that the training dataset plays a significant role in learning depth cues and this should be considered in situations where a connection with explainable results is required. 

This is a preliminary study, focusing on a single monocular depth cue, the relative size, and only a small set of methods. Explainability in depth estimation was considered on the basis of how humans estimate depth and a connection between the visual depth cues was attempted with the depth estimation provided by state-of-the-art approaches. To enable this, a new dataset was created, to mimic the original experiments in humans. This meant that the relative size cue should be isolated and no other depth cues should be present in the images. This study is ongoing and more datasets are created for each of the visual depth cues to assess the effectiveness of the existing methods, and thus, to conclude on the potential explainability in learning those cues. Overall, the final dataset will become a benchmark for testing the explainability of depth estimation methods.

\section{Conclusion}

In this study, we tried to approach explainability in depth estimation deep learning methods in terms of human perception. To this end, a specific visual depth cue was selected (the relative size) and a new dataset was created to mimic the experiments in humans. Three state-of-the-art pretrained methods were selected and tested against this dataset. As this dataset is limited to provide only a single cue, the accuracy of the methods indirectly reflects their success in learning the selected depth cue. In addition, it has been observed that the typical assessment metrics should be applied on rescaled versions of the image objects, in order to balance the estimated accuracy. Overall, the methods returned interesting accuracy results. Currently, we are expanding the dataset to include other visual depth cues and design new experiments to evaluate the efficiency of state-of-the-art methods.

%
%

%
%
%
%
\bibliographystyle{unsrt}
\bibliography{paper}
\end{document}